\documentclass[letterpaper, 10 pt, journal, twoside]{IEEEtran} 

\IEEEoverridecommandlockouts

\usepackage{multicol}
\usepackage{amsmath}
\usepackage{amssymb}
\usepackage{comment}
\usepackage{color}
\usepackage{graphicx}
\usepackage{mathtools}
\usepackage{subcaption}
\usepackage[hidelinks,bookmarks=true]{hyperref}

\newcommand{\argmax}{\operatornamewithlimits{argmax}}

\def\argmax{\mathop{\rm argmax}}

\def\sub{\mathop{\rm sub}}
\def\sup{\mathop{\rm sup}}

\def\atan2{\mathop{\rm atan2}}

\title{ 
An Analytical Lidar Sensor Model Based on Ray Path Information 
}

\author{Alexander Schaefer, Lukas Luft, Wolfram Burgard
\thanks{Alexander Schaefer and Lukas Luft contributed equally to this work.}
\thanks{This work has been partially supported by the European Commission in the Horizon
	2020 framework program under grant agreement 644227-Flourish; by the Graduate School
	of Robotics in Freiburg; by the State Graduate Funding Program of Baden-W\"{u}rttemberg.}
\thanks{All authors are with the Department of Computer Science, University of Freiburg, Germany.}
\thanks{\tt\small \{aschaef,luft,burgard\}@cs.uni-freiburg.de}}%

\author{Alexander Schaefer, Lukas Luft, and Wolfram Burgard%
\thanks{\copyright\ 2017 IEEE. Personal use of this material is permitted.  Permission from IEEE must be obtained for all other uses, in any current or future media, including reprinting/republishing this material for advertising or promotional purposes, creating new collective works, for resale or redistribution to servers or lists, or reuse of any copyrighted component of this work in other works.}
\thanks{Manuscript received: September 10, 2016; Revised February 6, 2017; Accepted January 25, 2017.}
\thanks{This paper was recommended for publication by Editor Francois Chaumette upon evaluation of the Associate Editor and Reviewers' comments. This work has been partially supported by the European Commission in the Horizon
2020 framework program under grant agreement 644227-Flourish; by the Graduate School
of Robotics in Freiburg; by the State Graduate Funding Program of Baden-W\"{u}rttemberg.}
\thanks{The first two authors contributed equally to this work. All authors are with the Department of Computer Science, University of Freiburg, Germany.}
\thanks{\tt\small \{aschaef,luft,burgard\}@cs.uni-freiburg.de}
\thanks{Digital Object Identifier (DOI): 10.1109/LRA.2017.2669376}}

\begin{document}

\maketitle


\begin{abstract}
Two core competencies of a mobile robot are to build 
a map of the environment and to estimate its own pose on the basis of this map and incoming sensor readings. 
To account for the uncertainties in this process, one typically employs probabilistic state estimation approaches combined with a model of the specific sensor.
Over the past years, lidar sensors have become a popular choice for mapping and localization.
However, many common lidar models perform poorly in unstructured, unpredictable environments, they lack a consistent physical model for both mapping and localization, 
and they do not exploit all the information the sensor provides, e.g. out-of-range measurements.
In this paper, we introduce a consistent physical model that can be applied to mapping as well as to localization.
It naturally deals with unstructured environments and makes use of both out-of-range measurements and information about the ray path.
The approach can be seen as a generalization of the well-established reflection model,
but in addition to counting ray reflections and traversals in a specific map cell, it considers the distances that all rays travel inside this cell. 
We prove that the resulting map maximizes the data likelihood and demonstrate that our model outperforms state-of-the-art sensor models in extensive real-world experiments.
\end{abstract}


\section{Introduction}
\IEEEPARstart{I}{n} the context of localization and mapping for mobile robots, sensor models serve two purposes:
First, the robot uses them to generate a map from recorded measurements;
second, they enable the robot to estimate its pose by relating subsequent sensor information to that map.

In practice, lidar sensors are widely used.
They send out laser rays and report how far they
travel before they are reflected by an object.
Ideally, the output distance reveals the closest object in a particular direction.
However, especially in unstructured 
outdoor environments with vegetation, two consecutive laser scans taken 
from the same point of view might return significantly different values.
The reason lies in the unpredictable interaction between the laser
ray and unstructured  objects, for example a tree canopy.
Ignorance about the thickness of single leafs,
their poses, etc., makes reflection a probabilistic process.

For lidar sensors, only a few probabilistic approaches formulate a consistent model for both mapping and localization with grid maps, e.g.
the reflection model \cite{Haehnel2003:dynamic_environments}
and related ray-tracing based approaches
\cite{Thrun2003:OccupancyMaps,Atanasov2014:Semantic,Leung2015:RFS,Yguel2008:GPUGridmapping}.
They tesselate the environment and assign to each voxel the probability that it reflects an incident laser ray.

\begin{figure}[t]
    \centering
\includegraphics[width=1\columnwidth]{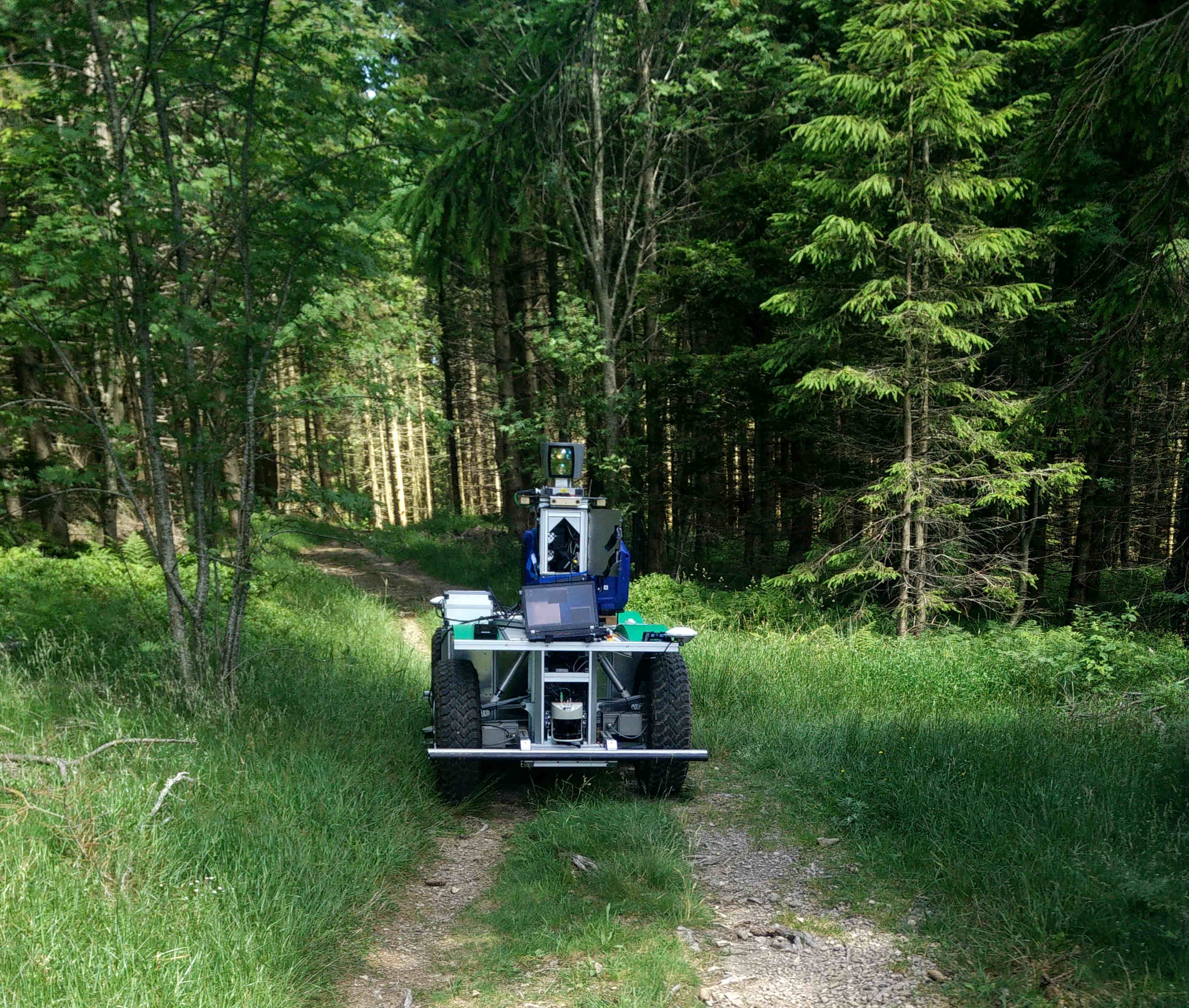}\\
\caption{Our mobile robot VIONA while recording the forest dataset.
}
\label{fig:title}
\end{figure}

\begin{figure*}[ht]
    \centering
\includegraphics[width=1.6\columnwidth]{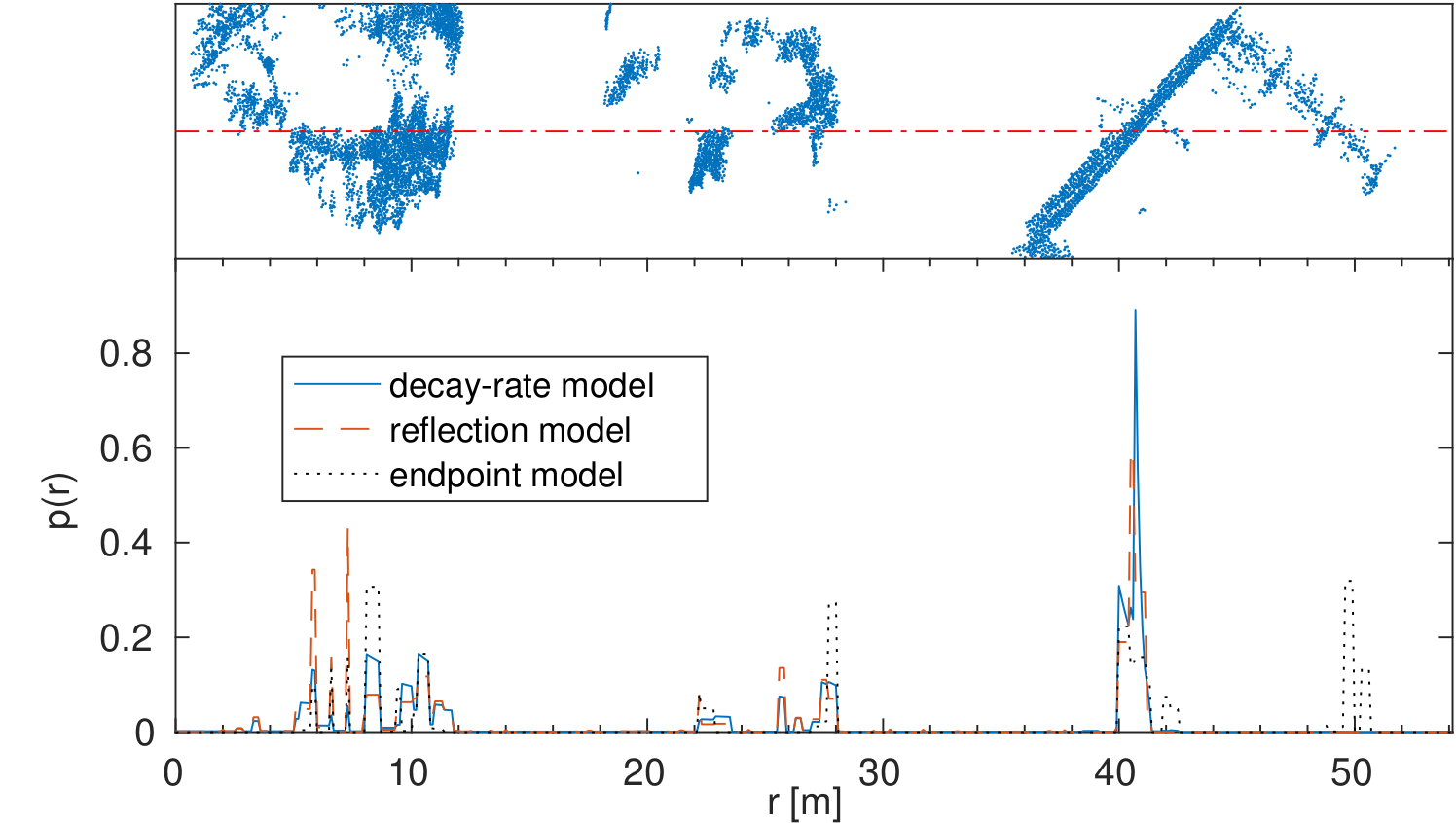}\\
\caption{The upper part of the image shows a section of the campus environment represented by a point cloud.
The dashed line represents a hypothetical laser beam traversing the scene from left to right.
It penetrates two treetops and a building.
The lower plot shows the corresponding measurement probabilities $p(r)$ obtained by the different sensor models.
The endpoint model attributes high probabilities to reflections at the right edge of the building ($r\approx 50$\,m)
because it ignores the ray trajectory and hence the wall at $r\approx 40$\,m.
In contrast, the two ray-casting based approaches attribute low probabilities to reflections behind the first wall.
The reflection model overestimates the probabilities in the treetops, as it does not account for the
distances the rays traveled within the treetop  voxels during the mapping process.
The overestimations of the endpoint model and the reflection model lead to lower relative probabilities at the
left wall of the building.
}
\label{fig:both}
\end{figure*}

In this paper, we introduce a novel probabilistic model for lidar sensors, which is a generalization of the aforementioned reflection model. 
In contrast to the latter, it relies upon a physical model of the interaction between the laser ray and the environment.
We model the probability that a ray traverses a specific region as an exponential decay process.
Based on the measurements collected during the mapping process, our sensor model assigns a decay rate to each point in space.
During the localization phase, we use this decay-rate map to determine the likelihood of incoming measurements.

This paper is structured as follows: Section~\ref{sec:rw} provides an overview over related work on lidar models. 
Section~\ref{sec:approach} describes how to build a decay-rate map from lidar measurements and how to compute the measurement likelihood for a given scan.
In Section~\ref{sec:details}, we prove that decay-rate maps maximize the likelihood of the underlying data and that our approach generalizes the reflection model.
Finally, Section~\ref{sec:evaluation} compares the performance of the proposed approach to state-of-the-art sensor models.

\section{Related Work}
\label{sec:rw}

In contrast to our concept, many other approaches address either mapping or localization.
Consequently, in the following section, we consider these two categories separately.

\subsection{Map Representations}

Occupancy grid maps, as introduced by Elfes~\cite{Elfes1989_OGM}, are widely used throughout the robotics community. 
They divide the environment into cells and assign to each of those a binary random variable that indicates whether
the cell contains an object.
A binary Bayes filter updates the distribution over these independent variables.
As opposed to our model, this approach assumes that the interaction between ray and map is deterministic:
The ray is reflected by the first occupied cell on its path.

Point clouds are a direct representation of the reflections measured by the lidar device.
However, they neglect out-of-range measurements and valuable information about the ray path.

Likelihood fields~\cite{Thrun2001} heuristically assign to each point in space the likelihood that a ray is reflected. 
Usually, this likelihood is derived from the distance to the nearest reflection observed during the mapping process.
This representation has the advantage that the likelihoods are functions of the space, which can be calculated in advance and stored in a distance map.
On the downside, it neglects the ray path information. As a consequence, the likelihood only depends on the endpoint and not on the objects along the ray.

Another popular map representation are reflection maps like used in \cite{Haehnel2003:dynamic_environments} and \cite{Thrun2003:OccupancyMaps}.
They assign to each cell a reflection probability,
which is determined by counting the rays that traverse the cell without reflection -- so-called misses -- and the rays that are reflected in the cell -- so-called hits.
Similar to our approach, reflection maps model the interaction between the beam and the map in a probabilistic way.
However, in addition to counting hits and misses, our approach considers the distances traveled within each cell.
The reflection model discards this information.

Instead of partitioning the map into a set of cubic voxels, Ferri et~al.~\cite{Ferri2015} use spherical voxels.
Bennewitz et~al.~\cite{Bennewitz2009:Reflection_Properties} explicitly handle erroneous measurements caused by the specific reflection properties of objects.
Ahtiainen et~al.~\cite{Ahtiainen2016:NDT_plus_permeability} use the reflection probability of a cell to decide whether it is traversable or not.

In contrast to grid-based approaches, feature-based maps describe the environment by a set of semantic objects.
The random finite set formulation as used in \cite{Atanasov2014:Semantic} and \cite{Leung2015:RFS} is a way to describe object detections.

There exist lots of other map representations that target specific applications. 
For example, Limosani et~al.~\cite{Limosani2015} use lidar in a long-term  mapping run in an office setting to model where dynamic objects like
humans are likely to be found.

\subsection{Sensor Models}

Sensor models can be divided into three categories: correlation-based, feature-based, and beam-based models \cite{Plagemann2007:GBP}.
Correlation-based models relate sensor readings to a given global map.
The popular endpoint model~\cite{Thrun2001}, for example, evaluates a likelihood field at the ray endpoints.
In this way, the endpoint model ignores information about the ray trajectory.
If both the global map and the local measurements are represented by point clouds, the iterative closest point method~\cite{Besl92:ICP}
or the normal distributions transform~\cite{Biber2003:NDT} can be used to determine the correlation
without the need for an explicit forward sensor model.
Feature-based approaches extract features from the sensor readings and compare them to the map.

Our model belongs to the class of beam-based approaches, which explicitly calculate the probability density of the distance measurement along the ray. 
As further instances of this class, \cite{Haehnel2003:dynamic_environments} reasons about dynamic objects,
and \cite{Thrun2003:OccupancyMaps} accounts for Gaussian sensor noise and false detections.
Thrun et~al.~\cite{Burgard2005:Probabilistic_Robotics} derive a basic beam-based model,
which De~Laet et~al.~\cite{DeLaet2008} augment by explicitly modeling and marginalizing dynamic objects.
Yguel et al.~\cite{Yguel2008:GPUGridmapping} address the problem that beam-based approaches are computationally expensive.
They present a GPU-accelerated mapping algorithm for several range sensors with different resolutions.
Mullane et al.~\cite{Mullane2009:LikelihoodFiltering} estimate the grid occupancy probabilities and
the corresponding detection likelihoods simultaneously rather than assuming a known measurement model.
In this way, their method accounts for false detections.

For a detailed survey on measurement models, see Chapter 12 in \cite{Mahler:Book}.

\section{Approach}
\label{sec:approach}

This section describes how to build a decay-rate map from lidar sensor readings and how to calculate the likelihood of a measurement using that map.
Table~\ref{tab:notation} provides an overview over the notation used throughout the paper.

\bgroup
\def\arraystretch{1.3}
\begin{table}[ht]
\centering
 \begin{tabular}
{ | r l | }
	\hline
  $i$ & voxel index \\
  $j$ & ray index \\
  $k(j)$		& index of voxel that reflects ray $j$ \\
  $v_i$			& $i^\textrm{th}$ voxel\\
  $\mathcal{I}$	& set of all voxels\\
  $d_i(j)$		& distance that ray $j$ travels inside $v_i$\\
  $H_i$			& total number of reflections in $v_i$\\
  $\lambda_i$	& decay rate in $v_i$ \\
  $\tau_i$		& mean ray length in $v_i$ \\
  $q_i$			& reflection probability in $v_i$ \\
  $s$			& sensor pose in map frame\\
  $m		$	& map \\
  $\lambda(x)$		& decay-rate map \\
  $r$			& measured ray length\\
  $x(r)$		& trajectory of ray for a fixed sensor pose\\ 
  $N(r)$		& probability that ray travels at least distance $r$\\
  \hline
\end{tabular}
\caption{Notation.}
\label{tab:notation}
\end{table}
\egroup

\subsection{The Basic Idea of the Decay-Rate Model}

The essence of our approach is to model the probability that a ray traverses a specific region as an exponential decay process. 
The decay rate of each point in the physical space is stored in a so-called decay-rate map.

To formalize this idea, we define $s$ as the sensor pose, which includes the origin and the direction of the ray, and $r$ as the distance between the sensor and the point of reflection.
For ease of notation, we write the measurement probability as
\begin{align}
\label{eqn:measurement_model}
p(r):=p(r\mid s, m).
\end{align}

In the present paper, we assume that the returned value $r$ is the actual distance
traveled by the beam, and model the relation between this distance and the map in a probabilistic fashion.
Measurement errors like Gaussian noise and false alarms are not in the scope of the proposed approach.
For approaches that account for these uncertainties, please see \cite{Mahler:Book}.

Under this assumption, the cumulative probability for a beam to travel at least
distance $r$ is
\begin{align}
\label{eqn:N}
N(r):=1-\int_0^r p(r')\ dr'.
\end{align}
For the measurement probability, it follows
\begin{equation}
\label{eqn:p_and_N}
 p(r)\stackrel{\eqref{eqn:N}}{=}-\frac{dN(r)}{dr}.
\end{equation}
Now, we introduce our essential idea: Locally, $N(r)$ obeys an exponential decay process:
\begin{align}
\label{eqn:differential_eqn}
\frac{dN(r)}{dr}=-\lambda\left(r\right) N\left(r\right).
\end{align}
This model is inspired by the following notion.
The physical space is filled with particles, and the probability that a laser ray
traverses a region in this space is proportional to the corresponding particle density.
Low densities correspond to permeable objects like bushes, while
high densities correspond to solid objects like walls.
For a ray that penetrates a region of constant particle density, $N(r)$
decreases exponentially over the traveled distance~$r$.

In our model, the decay rate $\lambda(x)$ is a property of the physical space.
We obtain $\lambda(r)=\lambda(x(r,s))$ by evaluating the decay rate along the trajectory of the ray.

Solving differential equation~\eqref{eqn:differential_eqn} for constant decay rate $\lambda$ yields
\begin{eqnarray}
\label{eqn:differential_eqn_solution_N}
 N(r) &\stackrel{\eqref{eqn:differential_eqn}}{=}& e^{-\lambda r}\\
 \label{eqn:differential_eqn_solution_p}
 p(r) &\stackrel{\eqref{eqn:p_and_N}+\eqref{eqn:differential_eqn_solution_N}}{=}& \lambda e^{-\lambda r},
\end{eqnarray}
assuming $N(0)=1$.
This solution is the basis of the mapping and localization algorithms derived in the following section.

\subsection{Mapping}
\label{sec:map}

For a given model, a map has to fully determine the interaction between a sensor  and the environment.
According to equation \eqref{eqn:differential_eqn}, the decay rate $\lambda$ meets this requirement.
Thus, we choose $\lambda(x)$ as map.
In order to relate the abstract parameter $\lambda$ to quantities which the sensor can observe,
we introduce $\tau$ -- the mean length which a ray travels in a hypothetical, infinitely large medium with constant $\lambda$,
before it is reflected:
\begin{align}
\label{eqn:tau}
\tau:=\mathbb{E}[r]= \int_0^{\infty}r\cdot p(r)\,dr \stackrel{\eqref{eqn:differential_eqn_solution_p}}{=} \lambda^{-1}.
\end{align}
On the basis of a finite number of measurements, the integral can be approximated as
\begin{equation}
\label{eqn:tau_approx}
 \tau = \lambda^{-1}\stackrel{\eqref{eqn:tau}}{\approx} H^{-1}\sum_{j\in\mathcal{J}} d(j),
\end{equation}
where $H$ is the number of recorded reflections, $\mathcal{J}$ is the set of measured rays,
and $d(j)$ is the distance that ray $j$ travels before it is reflected.
To determine $d(j)$, one uses ray tracing
between sensor position and reflection point.

To build a map of the environment, 
we tesselate the physical space using voxels $\left\{v_i\right\}_{i\in\mathcal{I}}$ of constant decay rates $\lambda_i$, so that
the decay-rate becomes a function of physical space:
\begin{equation}
\label{eqn:lambda_i_prel}
 \lambda(x\in v_i)= \lambda_i.
\end{equation}
Inspired by \eqref{eqn:tau_approx}, we define
\begin{equation}
\label{eqn:lambda_i}
 \lambda_i{:=}\frac{H_i}{\sum_{j\in\mathcal{J}} d_i(j)},
\end{equation}
where $H_i$ is the number of recorded reflections within $v_i$, 
and $d_i(j)$ is the distance that ray $j$ traveled within $v_i$.
With \eqref{eqn:lambda_i}, we can now determine our map -- the set $\{\lambda_i\}_{i\in \mathcal{I}}$ --
from sensor measurements. In practice, we have to account for finite memory.
Therefore, we compute $\lambda_i$ for all voxels inside a region of interest and assign a single prior to all points outside.

In Section~\ref{sec:ml}, we prove that the computation of the map parameters $\lambda_i$ according to \eqref{eqn:lambda_i} indeed maximizes the data likelihood.

\subsection{Localization}

During the localization phase, the robot uses the map to assign probabilities to measurements.
For a ray starting and ending in the same voxel $v_i$, \eqref{eqn:differential_eqn_solution_p} readily provides us with this probability.
Almost every ray, however, will traverse multiple voxels.
In order to determine the corresponding measurement probability,
we plug the piecewise constant decay rate as defined in~\eqref{eqn:lambda_i} into the differential equation~\eqref{eqn:differential_eqn} and solve for $N(r)$:
\begin{equation}
\label{eqn:N(r)}
 N(r)= \prod_{i\in\mathcal{I}} e^{-\lambda_i d_i}.
\end{equation}
To verify that \eqref{eqn:N(r)} satisfies the differential equation~\eqref{eqn:differential_eqn}, we need to differentiate $N(r)$ with respect to $r$.
Doing so, we need to keep in mind that for a particular $r$,
all but the last $d_i$ are constants obtained by ray tracing. Only the distance $d_k$ within the last voxel $v_k$ explicitly depends on $r$:
\begin{equation}
\label{eqn:d_k}
d_k= r - \sum_{ i\in\mathcal{I} \setminus \{k\} } d_i.
\end{equation}
With these prerequisites, the measurement likelihood becomes
\begin{equation}
\label{eq:p(r)}
 p(r)\stackrel{\eqref{eqn:p_and_N}}{=}-\frac{dN(r)}{dr}
 \stackrel{\eqref{eqn:N(r)}+\eqref{eqn:d_k}}{=}\lambda_k \prod_{i\in\mathcal{I}} e^{-\lambda_i d_i}.
\end{equation}

As described above, the values computed during mapping \eqref{eqn:lambda_i} and localization \eqref{eq:p(r)} are mainly linear combinations of values obtained by ray tracing.
Thus, the complexity of our method is determined by the complexity of the used ray tracing algorithm.
In particular, our approach has the same complexity as the reflection model, while it makes use of more measurement information.

\subsection{Integrating Out-of-Range Measurements}

Until now, we have implicitly assumed that the sensor always returns a real value $r$.
In practice, however, lidar sensors have a limited range $[r_{\min}; r_{\max}]$.
They return 
\begin{equation}
z := 
\begin{cases}
\sub & \textrm{for reflections below } r_{\min} \\
r & \textrm{for reflections in }[r_{\min}; r_{\max}] \\
\sup & \textrm{for reflections above } r_{\max}
\end{cases}
\end{equation}
Consequently, the measurement probability of a scan that contains $J$ rays becomes a mixture of probability densities and absolute probabilities:
\begin{align}
\begin{split}
\label{eqn:mixed_density}
p(z_{1},\ldots,z_{J}\mid s_1,\ldots,s_J,m)
\end{split}\\
\begin{split}
\nonumber \sim & \prod\displaylimits_{j\in\mathcal{J}_{\sub}} P(\sub\mid s_j,m)\\
&\cdot \prod\displaylimits_{j\in\mathcal{J}_\mathbb{R}} p(r_j\mid s_j,m) \\
&\cdot \prod\displaylimits_{j\in\mathcal{J}_{\sup}} P(\sup\mid s_j,m),
\end{split}
\end{align}
where $\mathcal{J}_{\sub}$, $\mathcal{J}_{\mathbb{R}}$, and $\mathcal{J}_{\sup}$ are the sets of ray indices that correspond to $z_j=\sub$, $z_j=r_j$, and $z_j=\sup$, respectively.

To compute the probabilities of out-of-range measurements, we integrate over all real values which they represent:
\begin{align}
P(\sub\mid s_j,m)	&=&	\int_0^{r_{\min}}p(r\mid s_j,m)\ dr \\
\nonumber P(\sup\mid s_j,m)	&=&	\int_{r_{\max}}^\infty p(r\mid s_j,m)\ dr,
\end{align}
with $p(r\mid s_j,m)$ as in \eqref{eq:p(r)}.

In the context of localization, the fact that \eqref{eqn:mixed_density} represents a mixture of probability densities and absolute probabilities does not bother us,
as we are typically interested in the relative probabilities between pose hypotheses. To obtain absolute probabilities, the measurement likelihood provided by~\eqref{eqn:mixed_density} has to be normalized.

\section{Mathematical Details}
\label{sec:details}

This section proves that the proposed mapping algorithm maximizes the data likelihood and derives the reflection model from our more general approach. 

\subsection{Decay-Rate Maps Maximize the Data Likelihood}
\label{sec:ml}

We prove that the map parameters $\lambda_i$ according to \eqref{eqn:lambda_i} maximize the likelihood of the underlying data by solving the following optimization problem:
\begin{eqnarray}
 m^*	&=&\underset{m=\{\lambda_i\}_{i\in\mathcal{I}}}{\argmax}	p\left( r_1,\dots,r_J \mid  s_1,\dots,s_J,m \right)\\
\nonumber	&=&\underset{m}{\argmax} 	\prod_{j\in\mathcal{J}} p\left( r_j\mid  s_j, m \right)\\
\nonumber		&=&\underset{m}{\argmax} 	\sum_{j\in\mathcal{J}} \log p\left( r_j\mid  s_j, m \right)\\
\nonumber		&\stackrel{\eqref{eq:p(r)}}{=}&\underset{m}{\argmax} 	\sum_{j\in\mathcal{J}} \log \left(\lambda_{k(j)} \prod_{i\in\mathcal{I}} e^{-\lambda_i d_i(j)}\right)\\
\nonumber		&=&\underset{m}{\argmax} 	\underbrace{\sum_{j\in\mathcal{J}} \left(\log\left(\lambda_{k(j)}\right) - \sum_{i\in\mathcal{I}} \lambda_i d_i(j)\right)}_{=:f(m)}.
\end{eqnarray}
With
\begin{align}
 \frac{\partial \log\left(\lambda_{k(j)}\right)}{\partial \lambda_i}=
  \begin{cases*}
      \frac{1}{\lambda_i}	& if $k(j)=i$ \\
      0        			& otherwise
  \end{cases*}
\end{align}
we obtain the partial derivatives of $f$:
\begin{equation}
\label{eq:f_part}
 \frac{\partial f(m)}{\partial \lambda_i} 
= \frac{H_i}{\lambda_i} - \sum_{j\in\mathcal{J}} d_i(j).
\end{equation}
Equation \eqref{eqn:lambda_i} satisfies both the necessary condition for $m^*$
\begin{equation}
\frac{\partial f(m)}{\partial \lambda_i}=0\ \forall\ i \in \mathcal{I}
\end{equation} 
and the sufficient condition
\begin{align}
\frac{\partial^2 f(\lambda)}{{\partial \lambda_i}^2}=-\frac{H_i}{{\lambda_i}^2} < 0.
\end{align}
Hence, the decay-rate map computed according to \eqref{eqn:lambda_i} is the most probable map given the sensor data.

\subsection{The Decay-Rate Model Generalizes the Reflection Model}
\label{sec:reflection_maps}

In order to show that the decay-rate model is a ge\-ne\-ra\-li\-za\-tion of the reflection model, we derive the latter from our approach with additional restrictive assumptions.
Reflection maps assign to each cell of the physical space a reflection probability 
\begin{equation}
\label{eqn:q_i}
q_i=\frac{H_i}{H_i+M_i},
\end{equation}
where $H_i$ is the number of reflections in $v_i$ recorded during mapping,
and where $M_i$ is the number of rays that penetrated $v_i$.
The model states that the probability of a ray ending in $v_k$ is
\begin{equation}
\label{eqn:p(k)}
 P\left( x(r) \in v_k\mid s,m\right)=q_k\prod_{i\in \mathcal{B}(k)}\left(1-q_i\right),
\end{equation}
where $\mathcal{B}(k)$ denotes the indices of the voxels through which the ray travels.


The first assumption inherent to the reflection model is that every ray travels the same distance in every voxel it traverses:
\begin{equation}
\label{eq:naive_assumption_1}
 d_i=  \begin{cases*}
      d		& if $i\in\mathcal{B}$ \\
      0        	& if $i\notin\mathcal{B}$
  \end{cases*}
\end{equation}
Applying this assumption to the decay-rate model, we obtain
\begin{equation}
\label{lambda_i_naive}
 \lambda_i	\stackrel{\eqref{eqn:lambda_i}}{=}	\frac{H_i}{\sum_{j\in\mathcal{J}} d_i(j)}\stackrel{\eqref{eqn:q_i}+\eqref{eq:naive_assumption_1}}{=}\frac{q_i}{d}.
\end{equation}
With this simplified version of $\lambda_i$, we derive the reflection probability~\eqref{eqn:p(k)} from our model:
\begin{eqnarray}
 P( x(r) \in v_k\mid s,m)	&=&							\int\displaylimits_{x(r)\in v_k} p(r\mid s,m)dr\\
\nonumber    &\stackrel{\eqref{eq:p(r)}}{=}&				\int\displaylimits_{ x(r)\in v_k} 	\lambda_k \prod_{i\in\mathcal{I}} e^{-\lambda_i d_i}dr\\
\nonumber    &\stackrel{\eqref{lambda_i_naive}}{=}&		\int\displaylimits_{ x(r)\in v_k} \frac{q_k}{d} \prod_{i\in\mathcal{B}} e^{-q_i}dr\\
\nonumber    &\stackrel{\eqref{eq:naive_assumption_1}}{=}&		q_k \prod_{i\in\mathcal{B}} e^{-q_i}\\
\nonumber    &{\approx}&		q_k\prod_{i\in \mathcal{B}}\left(1-q_i\right).
\end{eqnarray}
The second simplification implicitly made by the reflection model expresses itself in the transition between the last two lines:
The model aborts the Taylor series of the exponential after the first derivative.

We just argued that the standard reflection model can be seen as a special case of the decay-rate model.
Another way of looking at the relation between the two models is that the decay-rate approach
is formally equivalent to the standard approach in a grid where each cell is partitioned into subvoxels with constant $q_i$ within the original grid cell $v_i$.
This can be seen as follows.
The length of a ray that travels through the grid can be expressed by the number of traveled subvoxels $n$
and the subvoxel size $l$ as $r=n\,l$.
We get the cumulative distribution
\begin{equation}
N(r)=(1-q_i)^{n\,l}=e^{\log(1-q_i)n\,l},
\end{equation}
which obeys an exponential decay and has the same form as the decay-rate model \eqref{eqn:differential_eqn_solution_N}.
 Thus, one can formally switch from the decay-rate model to a fine-grained version of the reflection model by
 choosing the values $\{q_i\}$ such that \mbox{$\log(1-q_i)=-\lambda_i$}.

\section{Experiments}
\label{sec:evaluation}
\begin{figure*}	
	\centering
	\begin{subfigure}[t]{0.48\textwidth}
		\centering
		\includegraphics[height=3.65in]{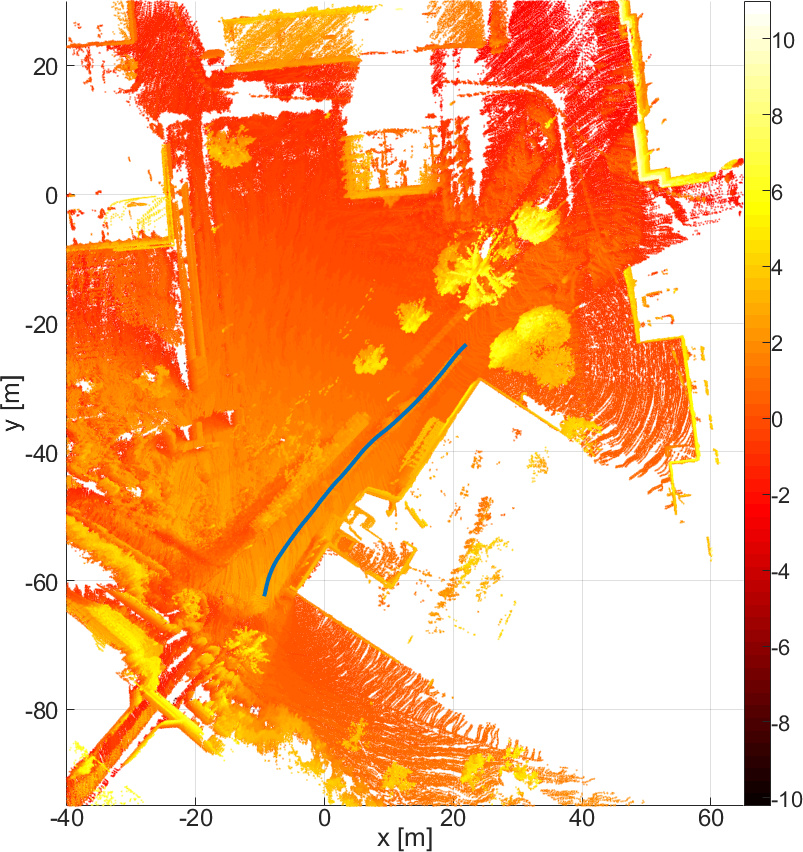}
		\caption{A section of the point cloud built from the campus dataset. The point heights are color-coded; the colorbar on the right tells which color denotes which height above the start position of the robot in [m].}
		\label{fig:pcmap_campus}		
	\end{subfigure}
	\quad
	\begin{subfigure}[t]{0.48\textwidth}
		\centering
		\includegraphics[height=3.65in]{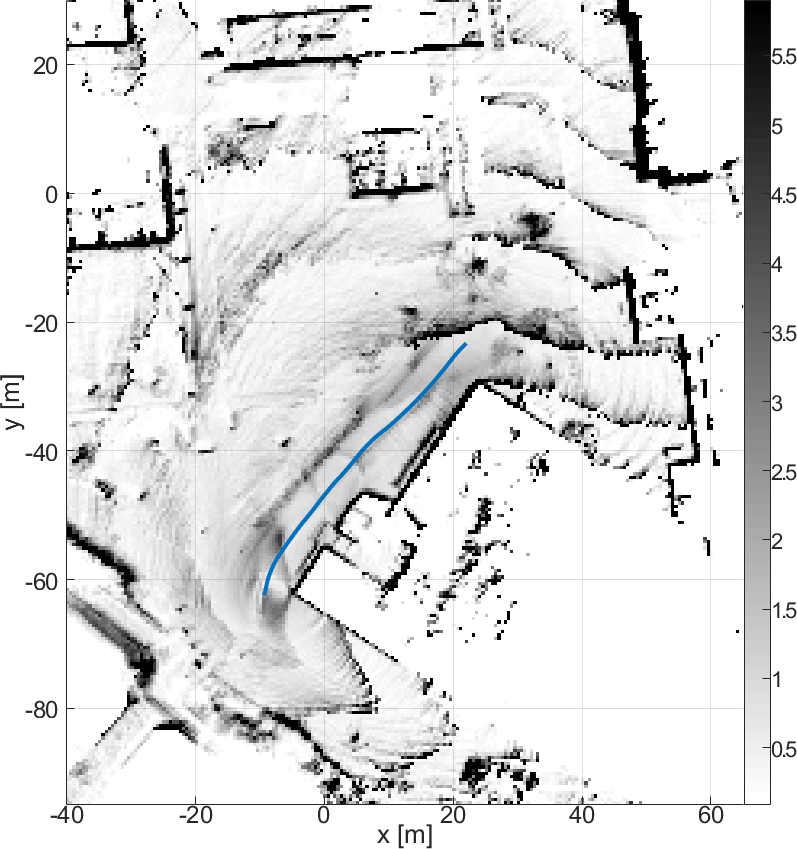}
		\caption{A section of the decay-rate map built from the campus dataset. This projection of the 3D decay-rate map onto the $x$-$y$ plane is computed by summing up the decay rates in $z$-direction. The colorbar on the right shows which color denotes which decay rate in [1/m]. Note that tree trunks are assigned a high decay rate, whereas the canopies have lower decay rates, e.g. at $(20,-10).$}
		\label{fig:decaymap_campus}
	\end{subfigure}
	\caption{Bird's eye view of a section of the campus dataset. The blue curve shows the robot trajectory ground truth as recorded by the Applanix localization system. The robot travels along a footpath that is framed by a small lawn with trees and bushes on the left and by a building on the right.}

\label{fig:pc}
\end{figure*}

In order to evaluate the proposed approach, we conduct extensive real-world experiments.
The data processed in these experiments were collected with the mobile off-road robot VIONA by Robot Makers, equipped with a Velodyne HDL-64E lidar sensor and an Applanix POS LV localization system.
We use the Applanix system, which fuses information coming from multiple GPS receivers, an IMU, and odometry sensors, as highly accurate pose ground truth.
The datasets were recorded in three different environments:
on the campus of the University of Freiburg, on a small trail in the middle of a forest, and in a park. All scenarios contain pedestrians.
The length of the recorded trajectories varied between 50\,m and 400\,m.
Figure~\ref{fig:title} shows the robot while recording the forest dataset.
Figure~\ref{fig:pc} shows the point clouds of the three datasets.

In the experiments, we compare the decay-rate model to two well-established, state-of-the-art sensor models: the reflection model~\cite{Haehnel2003:dynamic_environments} and the endpoint model~\cite{Thrun2001}.
For an illustration of the differences between these models, see Figure~\ref{fig:both}.

The set of measurements for mapping and the set for localization are disjoint.
We use the pose ground truth to perform mapping with known poses
for the different environments.
While our approach is applicable to any tesselation, in our experiments, we build maps consisting of cubic axis-aligned voxels with an edge length of 0.5\,m.
This way, the campus maps contain $444\times 406\times 43$ voxels, the park maps contain $515\times 561\times 41$ voxels, and the forest maps contain $393\times 403\times 86$ voxels.

\subsection{Monte Carlo Localization}
One of the main applications for sensor models is mobile robot localization.
In order to compare the different models with respect to localization accuracy in six dimensions, we run separate, identically parameterized particle filters for the three environments.
The filters only differ in the measurement models used to weight the particles in the correction step: The first filter employs the decay-rate model proposed in this paper, the second employs the reflectivity model, and the third employs the endpoint model.

We use 300 particles sampled from a Gaussian distribution with a variance of 1\,m in the horizontal plane, 0.2\,m vertically, and 0.1\,rad in every rotational dimension. The offset between the mean particle pose in the initialization step and the ground-truth start pose is sampled from this distribution, too.

To compare the robustness of the models, we also simulated sensor failures in the campus dataset by setting $10\%$ of the measurements to the minimum sensor range.

Table~\ref{tab:pf} shows the resulting averaged Euclidean distances between estimated and true poses for all recorded datasets and for the campus dataset with simulated sensor failures.

\begin{table}	
	\centering
    \begin{tabular}{ l | r r r }
    	& decay-rate model	& reflection model	&	endpoint model	\\ \hline
    campus 	& $\mathbf{0.230}$	& $0.284$	&	$0.280$\\
    campus*	& $\mathbf{0.252}$	& $0.284$	&	$0.366$\\
    forest 	& $\mathbf{0.331}$	& $0.352$	&	$0.417$\\
    park	& $\mathbf{0.088}$	& $0.089$	&	$0.124$
    \end{tabular}

	\caption{Particle filter estimation errors as Euclidean distances between ground truth and estimated position in [m], averaged over time.
	The scenario campus* includes simulated sensor failures. The best result for each scenario is highlighted in bold print.}
	\label{tab:pf}
\end{table}

\subsection{Evaluation of the Pose Likelihood}

To evaluate the measurement models independently of filter design, we employ two metrics that assess how well the pose likelihood derived from the output of the models matches ground truth.
First, we use the Kullback-Leibler divergence $\mathcal{D}\left(g\| h\right)$ to relate the pose likelihood $h$ to the ground truth $g$, which we approximate as a Dirac distribution.
With $z = \{z_1,\ldots,z_J\}$ and $s= \{s_1,\ldots,s_J\}$, we state:
\begin{eqnarray}
\label{eqn:KL}
\mathcal{D}\left(g\| h\right)	&=& \int g(s')\log\left( \frac{g(s')}{h(s')} \right)ds'\\
\nonumber				&=& \int \delta({s'-s)}\log\left( \frac{\delta({s'-s)}}{p(s'\mid m,z)} \right)ds'\\
\nonumber				&=& -\log\left[p(s\mid z,m)\right] + \eta  \\
\nonumber				&=& -\log\left[p(z\mid s,m)\right] + \eta' \\
\nonumber				&=& -\sum_{j=1}^J \log\left( p(z_j\mid s_j,m)\right) + \eta' \nonumber\\
\nonumber				&=:& \mathcal{D}'\left(g\| h\right) + \eta'.
\end{eqnarray}
In the evaluation, we omit the constant factor $\eta'$, as it is independent of the sensor model.
$\mathcal{D}'\left(g\| h\right)$ rewards high likelihoods at the real robot position, but it does not punish high likelihoods far from the real position.
To account for these false positives, we also employ the inverse Kullback-Leibler divergence
\begin{align}
\label{eqn:KLI}
		\mathcal{D}\left(h\| g\right)	&=& \int p(s'\mid m,z)\log\left( \frac{p(s'\mid m,z)}{\mathcal{N}\left(s'; s, \Sigma\right)} \right)ds' \\
\nonumber&\approx& \sum_{i=1}^M p(s_i\mid m,z)\log\left( \frac{p(s_i\mid m,z)}{\mathcal{N}\left(s_i; s, \Sigma\right)} \right).
\end{align}
To approximate the integral, we sum over $M$ poses $s_i$ sampled from a uniform distribution in a circular area centered at the true pose $s$.
We then obtain $p(s_i\mid m,z)$ by normalizing $p(z\mid s_i,m)$ over all $s_i$ and assume the real position to be distributed according to
$\mathcal{N}\left(s'; s, \Sigma\right)$.
Plagemann et~al.\ \cite{Plagemann2007:GBP} use a similar metric.
Table \ref{tab:both_KL} shows the corresponding results.

\bgroup
\def\arraystretch{1.4}
\begin{table*}	
	\centering
	\begin{subtable}[t]{0.48\textwidth}
		\centering
    \begin{tabular}{ l | r r r }
    	& decay-rate model	& reflection model	&	endpoint model	\\ \hline
    campus 		& $\mathbf{6.07 \cdot 10^4}$	& $6.99 \cdot 10^4$	&	$1.01 \cdot 10^5$ \\
    forest 		& $\mathbf{2.70 \cdot 10^4}$	& $3.33 \cdot 10^4$	&	$5.02 \cdot 10^4$ \\
    park		  & $\mathbf{1.11 \cdot 10^8}$	& $1.14 \cdot 10^9$	&	$1.16 \cdot 10^9$ \\ 
    \end{tabular}
		\caption{Divergence $\mathcal{D}'\left(g\| h\right)$ between Dirac-distributed ground truth and pose likelihood as defined in \eqref{eqn:KL}.
		Low values indicate high pose likelihoods at the true position. 
		The values are computed over all measurements in the dataset.}
	\label{tab:KL}
	\end{subtable}
	\quad
	\begin{subtable}[t]{0.48\textwidth}
		\centering
    \begin{tabular}{ l | r r r }
    	& decay-rate model	& reflection model &	endpoint model \\ \hline
    campus 		& $\mathbf{1.87}$	& $4.44$	&	$2.09$ \\ 
    forest 		& $\mathbf{0.96}$	& $1.41$	&	$1.14$ \\ 
    park		  & $\mathbf{3.56}$	& $4.64$	&	$4.17$ \\ 
    \end{tabular}
		\caption{Inverse Kullback-Leibler divergence $\mathcal{D}\left(h\| g\right)$ between Gauss-distributed ground truth and pose likelihood as in \eqref{eqn:KLI}, averaged over all scans.
		Low values indicate low pose likelihood far away from the true pose.
		We used M=50 samples from a uniform distribution within a circular area with radius 2.5\,m centered at the true robot pose.}
	\label{tab:KL_inverse}
	\end{subtable}
	\caption{Kullback-Leibler divergence between ground truth distribution and pose likelihood for different sensor models.
	For both metrics, smaller numbers correspond to higher similarity to ground truth. The best results are printed in bold.}
	\label{tab:both_KL}
\end{table*}
\egroup

Note that it is impossible to directly compare the output of the three models,
as one model returns absolute probabilities, the other probability densities,
and yet another heuristic values. To account for that, we always convert the output for real-valued measurements to probability densities
and the output for out-of-range measurements to absolute probabilities, as described in the following.

The reflection model yields absolute probabilities for both real-valued and out-of-range measurements.
For the former, we assume an underlying density that is implicitly integrated over the voxel that reflects the ray:
\begin{equation}
 P\left(x(r)\in v_k\right)	=	\int_{r\mid x(r')\in v_i} p(r'\mid s,m)dr'.
\end{equation}
As all rays ending in one voxel have the same probability, we conclude
\begin{align}
P\left(x(r)\in v_k\right)=p(r\mid s,m)\int_{r'\mid x(r')\in v_i} dr'.
\end{align}
Now we can identify the underlying probability density
\begin{align}
p(r\mid s,m)={P\left(x(r)\in v_k\right)}\left(\int_{r'\mid x(r')\in v_i} dr'\right)^{-1}.
\end{align}

The endpoint model assumes an absolute probability $P$ as prior for out-of-range measurements.
For measurements within the sensor range, it outputs heuristic values. 
To obtain the corresponding probability density, for each ray, we normalize the integral over all values within the sensor range to $1-P$.

As the decay-rate model already expresses the probabilities as required, all models are now comparable to one another.

\subsection{Discussion of Results}

The results of the localization experiments are listed in Table \ref{tab:pf}.
The proposed decay-rate model outperforms the two standard approaches on all datasets.
This is due to the fact that the decay-rate model leverages more of the information the sensor provides.

In the campus environment, the endpoint model performs better than the reflection model.
In the other, less structured environments, and in the scenario with sensor failures, the reflection model outperforms
the endpoint model.
We attribute this to the fact that especially in unstructured environments, the ray path information is more informative than the distance to the nearest point.

A comparison of the results of the campus dataset with campus*, which contains simulated sensor failures, indicates that
the two beam-based approaches are more robust against outliers than the endpoint model.

Although we chose a poor initial estimate, the particle filter converges to the true position for all datasets and all models.
The park dataset is recorded over the longest period of time.
Therefore, the bad initialization has less impact in this scenario than in the other three.

The evaluation of the pose likelihoods are listed in Table \ref{tab:both_KL}.
These results are more informative than the particle filtering results, as the latter are influenced by parameters and design choices.
The proposed decay-rate model outperforms the two baseline approaches in all scenarios:
In contrast to the endpoint model, it leverages ray-path information, and in contrast to the reflection model, it considers the distances the rays traveled within the cells.

\section{Conclusion and Future Work} 
\label{sec:conclusion}

In this paper, we introduce a physics-inspired, probabilistic lidar sensor model. 
As a generalization of the reflection model, it can consistently be applied to both mapping and localization.
We prove that the resulting maps maximize the data likelihood.
In extensive experiments, our model outperforms state-of-the-art measurement models in terms of accuracy.

Our approach models the uncertainty in the interaction between a ray and the environment.
In the future, we will extend it to account for additional measurement uncertainties like Gaussian noise and false detections.
Currently, we are working on a GPU-accelerated, real-time capable implementation on our off-road robot
and plan to build a SLAM framework based on the proposed approach.
We will evaluate the lidar calibration performance using ground-truth data obtained by SLAM, and we will also investigate different front-ends and methods for data association.
In this context, we plan to benchmark the localization accuracy and the computational requirements of all three sensor models.
We are also working on a differentiable extension of our model.


\bibliographystyle{IEEEtran}
\bibliography{IEEEabrv,decay}

\end{document}